\documentclass[runningheads]{llncs}

\usepackage{eccv}

\usepackage{eccvabbrv}

\usepackage{graphicx}
\usepackage{booktabs}

\usepackage[accsupp]{axessibility}  

\usepackage{hyperref}

\usepackage{orcidlink}

\usepackage{amsmath}
\usepackage{amssymb}
\usepackage{enumitem}
\usepackage{colortbl}
\usepackage{multirow}
\usepackage{overpic}
\usepackage{microtype}
\definecolor{mygray}{rgb}{0.9000,0.9000,0.9000}
\newcommand{\ourdataset}{WildBe\xspace}

\definecolor{novelcolor}{rgb}{0.8, 1.0, 0.9}

\usepackage[capitalize]{cleveref}
\crefname{section}{Sec.}{Secs.}
\Crefname{section}{Section}{Sections}
\Crefname{table}{Table}{Tables}
\crefname{table}{Tab.}{Tabs.}

\begin{document}

\title{{\color{teal}Wild Be}rry image dataset collected in Finnish forests and peatlands using drones}

\titlerunning{\textbf{Wild Be}rry image dataset}

\author{Luigi Riz\inst{1}\orcidlink{0009-0000-4687-4297} \and
Sergio Povoli\inst{1}\orcidlink{0009-0006-2227-6931} \and
Andrea Caraffa\inst{1}\orcidlink{0009-0004-7775-924X
} \and
Davide Boscaini\inst{1}\orcidlink{0000-0003-4887-2038} \and
Mohamed Lamine Mekhalfi\inst{1}\orcidlink{0000-0002-4295-0974} \and
Paul Chippendale\inst{1}\orcidlink{0000-0002-2471-0424} \and
Marjut Turtiainen\inst{2} \and
Birgitta Partanen\inst{2}\orcidlink{0000-0003-4860-2142} \and
Laura Smith Ballester\inst{3}\orcidlink{0000-0002-9855-4655} \and
Francisco Blanes Noguera\inst{3}\orcidlink{0000-0002-9234-5377} \and
Alessio Franchi\inst{4}\orcidlink{0000-0001-8840-3012} \and
Elisa Castelli\inst{4} \and
Giacomo Piccinini\inst{4}\orcidlink{0000-0003-2807-0871} \and
Luca Marchesotti\inst{4} \and
Micael Santos Couceiro\inst{5}\orcidlink{0000-0001-6641-6090} \and
Fabio Poiesi\inst{1}\orcidlink{0000-0002-9769-1279}
}

\authorrunning{L.~Riz \and S.~Povoli et al.}

\institute{Fondazione Bruno Kessler, Italy \and
Arctic Flavours Association, Finland \and
Universitat Politècnica de València, Spain \and
GemmoAI, Ireland \and
Ingeniarius, Portugal
}

\maketitle


\begin{abstract}
Berry picking has long-standing traditions in Finland, yet it is challenging and can potentially be dangerous. 
The integration of drones equipped with advanced imaging techniques represents a transformative leap forward, optimising harvests and promising sustainable practices. 
We propose \ourdataset, the first image dataset of wild berries captured in peatlands and under the canopy of Finnish forests using drones. 
Unlike previous and related datasets, \ourdataset includes new varieties of berries, such as bilberries, cloudberries, lingonberries, and crowberries, captured under severe light variations and in cluttered environments. 
\ourdataset features 3,516 images, including a total of 18,336 annotated bounding boxes.
We carry out a comprehensive analysis of \ourdataset using six popular object detectors, assessing their effectiveness in berry detection across different forest regions and camera types.
\ourdataset is publicly available on HuggingFace (\href{https://huggingface.co/datasets/FBK-TeV/WildBe}{https://huggingface.co/datasets/FBK-TeV/WildBe}).
\end{abstract}
\section{Introduction}\label{sec:introduction}

Berry picking is a diffused activity in Finland, engaging 54\% of Finnish households in 2011. 
This practice not only reflects cultural significance but also contributes to the economy.
The collective harvest of wild berries for home use, specifically lingonberries, bilberries, and cloudberries (also known as wild blueberries), is 4.1 kg, 4.9 kg, and 0.5 kg per household in 2011 on average, respectively \cite{Vaara2013}.
This results in a significant annual yield, with bilberries alone accounting for 184 million kilograms during an average berry year \cite{Turtiainen2007}, underscoring the potential for substantial economic and nutritional benefits.
Cloudberries, lingonberries, and bilberries are highly valued in their natural state. 
Bilberries, in particular, are noted for their potential health benefits, including anti-inflammatory properties and the ability to address conditions such as hyperglycemia, cardiovascular disease, cancer, diabetes, dementia, and other age-related ailments \cite{Chu2011biblerry}.
Despite the economic opportunities presented by wild berry harvesting, the industry faces inherent challenges. 
Earnings for pickers are tied to the type, quality, and weight of berries collected, often compelling them to work long hours without breaks.
This highlights the need for balance between maximising economic returns and ensuring fair labour practices.

\begin{figure}[t]
    \centering
    \includegraphics[width=0.9\columnwidth]{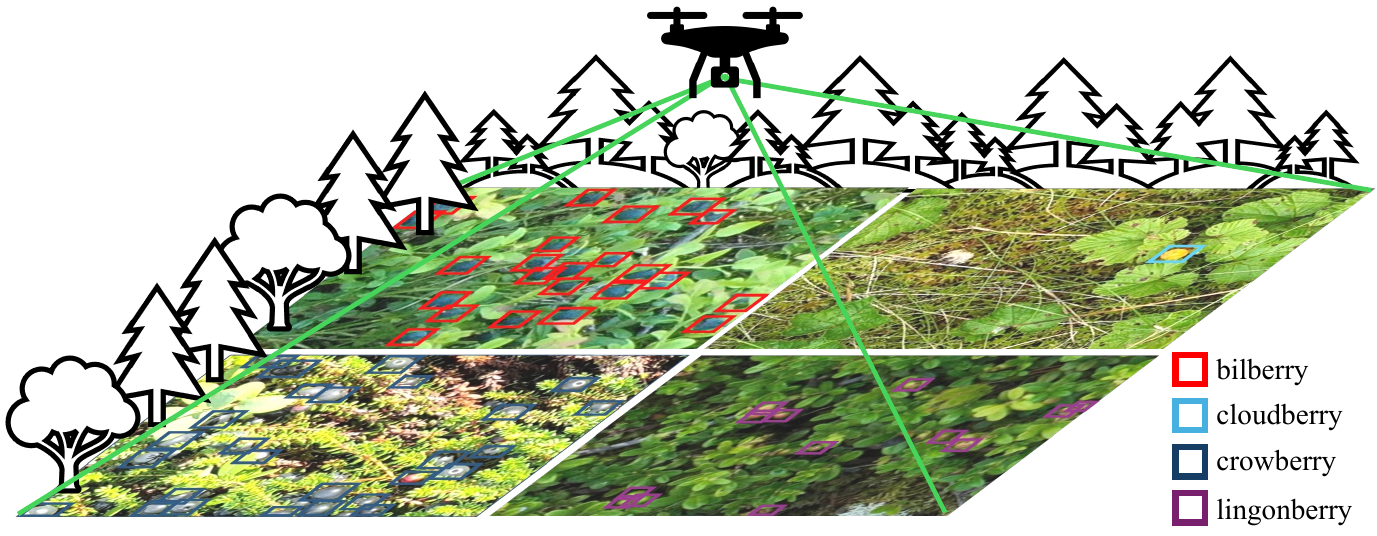}
    \caption{The \ourdataset dataset comprises images of wild berries, including bilberries, cloudberries, crowberries, and lingonberries, captured in peatlands and the undercanopy of Finnish forests using drones. Images are manually annotated with bounding boxes.
    }
    \label{fig:teaser}
\end{figure}

In the context of berry picking, the integration of drones equipped with advanced imaging techniques represents a transformative leap forward \cite{Akiva2020craid}.
The dynamic landscapes and the often unpredictable nature of wild berry habitats call for sophisticated tools to optimise harvests and ensure sustainable practices.
Drones, equipped with high-resolution cameras and multispectral imaging, are key to enable precise mapping of berry-rich areas, significantly reducing the time and physical effort required for berry pickers to locate good berry picking sites.
Imaging techniques can process the collected data enabling a more targeted and effective approach to berry picking.
Through the analysis of images, it is possible to assess the health and ripeness of berries, predict yield volumes, and even identify potential threats from pests or diseases. 
This information can also be valuable for managing the health of berry populations and their ecosystems. 

Berry detection in images presents several challenges.
Varying lighting conditions can significantly affect the berry visibility and colour, making it difficult to consistently identify and classify them across different times of the day or under changing weather conditions. 
Dense foliage and overlapping branches can obstruct views and lead to partial occlusions of the berries. 
Cluttered backgrounds, typical of forests, can lead to high false positive rates in berry detection algorithms, as models struggle to distinguish berries from similarly coloured objects.
The movement of the drone introduces motion blur and changes in perspective, which can further challenge the stability and accuracy of detection models.
In order to develop algorithms for berry detection, it is necessary to have large and diverse datasets of annotated images.
However, such datasets are difficult to obtain as collecting large amounts of images captured from drones in wild forests and peatlands, and annotating them is costly.
Hereafter, we use the term ``forests'' to refer to both forests and peatlands for simplicity.

In this paper, we propose \ourdataset, the first image dataset of wild berries that is collected in Finnish forests using drones.
To the best of our knowledge, the only dataset featuring wild berries is the CRAID dataset \cite{Akiva2020craid}.
CRAID consists of the largest collection of drone imagery from cranberry cultivation fields, gathered to train and evaluate the network for segmentation and counting of cranberries.
CRAID is composed of 21,436 cranberry images of resolution 456$\times$608 pixels, captured with a Phantom 4 drone.
Unlike CRAID, \ourdataset contains images of bilberries, cloudberries, crowberries, and lingonberries (Fig.~\ref{fig:teaser}).
\ourdataset is collected in forests, thus featuring several challenges like severe light variations, cluttered environments like tree branches, lichens, rocks, etc., and berries with different levels of ripeness.
We provide a detailed description of our dataset and perform a comprehensive experimental analysis using six popular object detectors, \ie, Faster R-CNN (2015)~\cite{ren2015faster}, 
VarifocalNet (2021)~\cite{zhang2021varifocalnet}, 
GLIP (2022)~\cite{li2022glip}, 
DINO (2023)~\cite{zhang2023dino}, 
ObjectBox (2022)~\cite{zand2022objectbox}, and 
YOLOv8 (2023)~\cite{Jocher2023YOLOv8}.
We experimentally evaluate these algorithms when trained and tested on mixed data, and assess their generalisation ability when trained and tested in different forest regions and with different sensors.

\section{Berries}

Cloudberry (\textit{Rubus chamaemorus}) is a plant species naturally found in boreal and arctic zones, although it also occurs in the mountainous regions of Central Europe.
Cloudberry plant's leaves are rounded with a toothed edge and have a wrinkled appearance. 
They typically grow in a pattern that forms a rosette at the base of the plant. 
Cloudberry flowers are small, white, and have five petals. 
The flowers grow alone rather than in clusters, emerging from the centre of the leaf rosette.
Cloudberries are amber-coloured berries that turn from red to soft golden-yellow or amber when ripe. Each berry is made up of multiple drupelets (similar to a raspberry or blackberry) and is about 1-2 cm in diameter. Cloudberry plants are relatively low to the ground, typically growing no more than 10-25 cm tall.

Lingonberry (\textit{Vaccinium vitis-idaea}) and bilberry (\textit{Vaccinium myrtillus}) have adapted to a wide range of different site and land types in coniferous ecosystems and, as a result, are widely distributed across Europe and northern Asia. 
In Finland, bilberry is typical and abundant, especially in conifer heath forests of medium site fertility (\eg, mesic heath forests).
Lingonberry is most typical in light pine-dominated dryish (sub-xeric) heath forests. 
Both species also occur and produce yields in marginal types of forest (\eg, fell forests), and on pristine and drained peatland sites \cite{Turtiainen2015}.
The leaves of the lingonberry plant are small, oval-shaped, and have smooth edges. 
They are dark green, glossy on the top, and can sometimes exhibit a slight reddish tint along the edges. 
The lingonberry plant retains its leaves, which even survive through the winter. 
Lingonberry flowers are bell-shaped, white to pale pink, and grow in small clusters. 
They are small and round, with a diameter of about 5-8 mm. 
Berries are initially light green, turning red upon ripening. 
Lingonberry plants are low-growing shrubs, typically reaching a height of 10-30 cm. 
The plants have a woody stem and can spread over a wide area.
Differently, the leaves of the bilberry plant are small, oval to elliptical, and have finely serrated edges. 
They are bright green and soft, growing along the slender, green branches. 
The flowers are small, usually pink in colour, and their shape is ball-like. 
The berries are round and small, about 5-8 mm in diameter. 
They are either dark blue and waxy or black and shiny.
Bilberry plants are low-growing shrubs, usually about 20-40 cm in height. They form dense, twiggy clumps and can spread over the ground in extensive patches, particularly in undisturbed habitats.
\section{Hardware}

\subsection{Multirotor drones}

In our pursuit of developing an intelligent solution to support berry pickers, under-canopy flying drones stand as a key tool, particularly lightweight drones (LWD). 
We select LWD based on key technical specifications that ensure performance in energy autonomy, sensing payload, communication technologies, integration with the Robot Operating System (ROS), autonomous operation, durability, and maneuverability \cite{quigley2009ros}. 
We capture data using commercial solutions such as DJI, specifically the Mini 2, Mavic 2 Pro, and Mavic 3M models. 
These models are renowned for their reliability and performance in agricultural tasks \cite{puri2017agriculture}. 
We also embark on designing our own drone, namely Scout v2, to comply with additional criteria established in the project, including ROS integration and autonomous navigation under the canopy. Although this custom-designed drone incorporates multiple stereo cameras to meet autonomous navigation requirements, we focus on data collected using the installed GoPro 11, as described in the next section.

\subsection{Cameras}

\ourdataset features different sensors, ranging from mobile, action, to drone cameras. 
Data also includes images captured with the Xiaomi 12X smartphone, equipped with a 50MP Sony IMX766 camera \cite{khatun2024comprehensive}. 
The DJI Mini 2 drone offers high-quality imagery with its 12MP camera, adaptable to various environmental conditions \cite{sorbelli2023yolo}. 
The Mavic 2 Pro mounts the Hasselblad L1D-20c camera with a 1-inch CMOS sensor, known to excel in low-light environments, thanks to its extended ISO range and enhanced dynamic range \cite{burdziakowski2021uav}. 
The DJI Mavic 3M mounts a 20MP RGB multispectral camera, already employed in agriculture applications \cite{sa2017weednet}. 
The GoPro 11, mounted on our custom drone named Scout v2, provides action-oriented imaging with its 27-megapixel sensor and HyperSmooth 5.0 video stabilisation technology. 
To the best of our knowledge, the GoPro 11 has not yet been utilised in agriculture applications; however, prior versions of the camera have demonstrated performance comparable to high-end models \cite{andritoiu2018agriculture}. 
Table \ref{tab_devices_dataset} summaries the cameras used.
\section{Dataset}\label{section_dataset}

\ourdataset was collected in the forests of Ilomantsi (Finland) in July 2023. 
It comprises 3,516 images, extracted from 555 raw images selected from different drone models and devices (Sec.~\ref{subsection_annotation_procedure}), and encompasses variations of cameras, albedos, forests, terrains, heights, perspective changes, and includes a total of 18,336 berries manually annotated as bounding boxes. 
Berry species include bilberries, cloudberries, crowberries, and lingonberries, the last at different ripening stages.
To provide a detailed reference of where the data was captured, \ourdataset includes metadata for each image, sourced from various devices (\cref{tab_devices_dataset}). 
This metadata includes: timestamp of each captured image, geographical coordinates, altitude, and camera details.
While DJI drones automatically capture this metadata, images from Xiaomi and GoPro cameras include only the capture date.

\begin{table}[t]
\centering
\caption{Devices used in \ourdataset data collection along with the resolution in pixel (width $\times$ height) of the corresponding images. 
\# images represent the image crops that we extracted from 555 raw images (the number of raw images is shown in parentheses).
}
\vspace{-2mm}
\tabcolsep 10pt
\label{tab_devices_dataset}
\begin{tabular}{@{}lccl@{}}
\toprule
Drone & Resolution [pixels] & metadata & \# images \\
\midrule
DJI Mavic 3M & $5280\times3956$ & \checkmark & 1345 (255) \\
DJI Mini 2 & $5280\times3956$ & \checkmark & 95 (23) \\
DJI Mavic 2 Pro & $5472\times3648$ & \checkmark & 1069 (80) \\
Xiaomi 12X & $2304\times4096$ & & 966 (182) \\
GoPro 11 & $5312\times2992$ & & 41 (15) \\
\bottomrule
\end{tabular}
\label{tab:sensors}
\end{table}

\subsection{Data collection requirements}\label{subsection_data_collection_requirements}

\ourdataset is aimed at berry detection tasks. 
The primary challenge lies in the small size of  berries, ranging from 5 to 8mm for bilberries and lingonberries, and can be as large as 20 mm for cloudberries. 
For accurate detection, it is essential that these berries span between 15 and 25 pixels. 
We establish this requirement based on a controlled experiment, which suggests that achieving a Ground Sampling Distance (GSD—metric distance between pixel centres) of 0.5mm is crucial. 
This calculation assumes an optimistic berry size of 10mm, aiming for a berry representation of 20 pixels in an image. 
The target GSD, in conjunction with the sensor specifications, indicates the required flight altitude of the drone above the berries to achieve optimal detection. 
For example, to maintain a desired GSD, the DJI Mavic 3M must operate at a height of approximately 1.65m above berries, translating to roughly 1.9m above ground level.
In practice, the flight height $h$ can be computed as
\begin{equation}\label{eq:height}\nonumber
    h = \frac{Res_h \cdot GSD}{2 \cdot tan(FOV_h/2)} = 
    \frac{5280 \cdot 5 \cdot 10^{-4} m}{2 \cdot tan(77.44^\circ/2)} = 1.65 m
\end{equation}
where $Res_h$ is the camera's horizontal resolution, and $FOV_h$ is the horizontal camera field of view.
Note that, given the assumption that berries are spherical, hence encapsulated within a square bounding box, applying the vertical camera specifications yields identical results.
Given the minimum berry size in pixels, the maximum permissible flight altitude would be 2.2m above the berries.

\subsection{Pre-processing and annotations}\label{subsection_annotation_procedure}

Given the original images captured by drones, we create smaller versions through cropping to enhance the efficiency of object detection algorithms. 
The cameras we use have different resolutions (\cref{tab_devices_dataset}). 
We select 555 original images from the collection, taking into account factors such as species, device, and lighting conditions, and post-process them into non-overlapping image crops of size 528$\times$396 pixels. 
In instances where image crops\footnote{Hereafter, we refer to these as `images' for simplicity.} are smaller, we rescale them using bilinear interpolation.
These image crops are manually annotated with bounding boxes in the YOLO format (\ie, 
($class, x_{center}, y_{center}, width, height$). 
The classes correspond to cloudberries, lingonberries, crowberries, and bilberries.
For each image, bounding boxes are drawn as tightly as possible around the berries to minimise the inclusion of background. 
Since bilberries and crowberries are both dark in colour and difficult to distinguish, annotators rely on the shape of the leaves to differentiate them (bilberry leaves are wide-open, whereas crowberry leaves are needle-shaped). 
\ourdataset also accounts for varying ripeness levels of lingonberries, from green to reddish hues, treating all as equivalent representations of the same species. 
Although less represented in numbers, cloudberries can be easily identified based by their distinctive yellow colour.

\begin{figure}[t]
\begin{center}
  \begin{tabular}{@{}c@{\,\,\,}c}
    \begin{overpic}[width=.45\columnwidth]{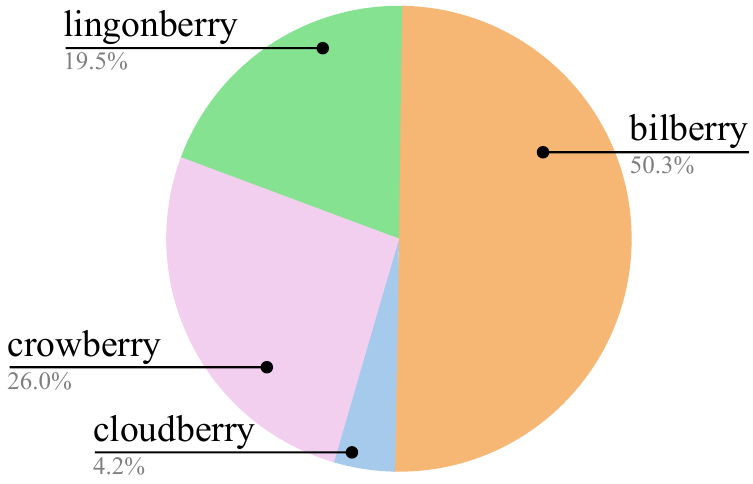}
    \end{overpic} &
    \begin{overpic}[width=.35\columnwidth]{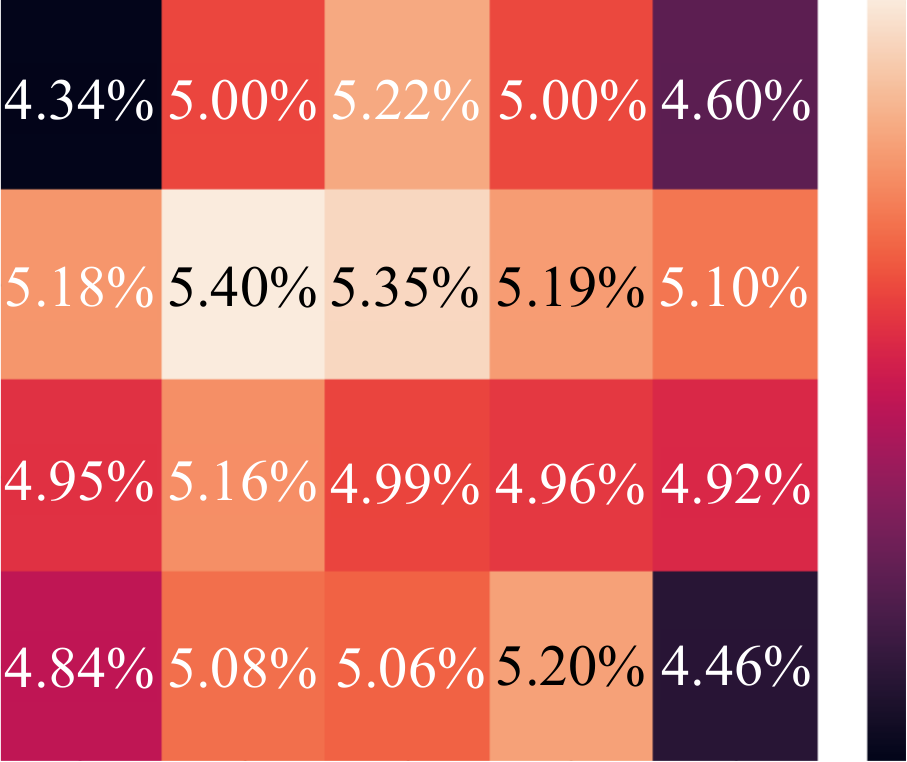}
    \end{overpic} \\
  \end{tabular}
\end{center}
\vspace{-5mm}
\caption{\ourdataset statistics: (left-hand side) proportion of annotations for each
class within the dataset and (right-hand side) distribution of bounding boxes across the images.}
\label{fig:pie_heat}
\end{figure}

\subsection{Annotation statistics}

\cref{fig:pie_heat} illustrates the proportion of annotations for each class within the dataset (on the left-hand side) and the distribution of bounding boxes across the images (on the right-hand side).
We can observe that the majority of the annotations are for bilberries, while cloudberries constitute the minority.
However, due to the distinctive yellow appearance of cloudberries, experiments demonstrate that detectors are particularly effective in identifying them.
Moreover, it is also evident that the annotations are evenly distributed across the images, which aids in preventing the development of biases during the training of algorithms.

\section{Experiments}

\subsection{Experimental setup}

We divide \ourdataset into disjoint splits for training, validation, and testing, containing 3164, 176, and 176 images, which correspond to 16.5K, 856, and 965 instances, respectively.
We conduct two sets of experiments to evaluate different object detection algorithms (hereafter referred to as algorithms).
In the first set of experiments, we label each bounding box with the generic class ``berry'', to obtain a general purpose ``berry detector'' that focuses on berry localisation.
We refer to this setting as \emph{single-class}.
In the second set of experiments, we set up a multi-class berry detection task to assess both the localisation and classification capabilities of the algorithms.
We refer to this setting as \emph{multi-class}.
We evaluate the performance of algorithms trained in the multi-class setting when applied to the single-class setting, specifically for assessing the ability to detect the presence of berries, regardless of their estimated class.
By exploiting \ourdataset metadata, we also subdivide the data into folds by the location of acquisition (four different areas) and by the device used to capture the images (five different cameras).
We evaluate the algorithms in the transfer learning setting, in which data coming from a single fold is left out during training and considered only at test time.
Lastly, we test the algorithms on the test split of the CRAID dataset~\cite{Akiva2020craid} when they are trained on the single-class \ourdataset, in order to evaluate their cross-dataset generalisation capabilities.
We use COCO evaluation and report results in terms of Average Precision (AP)~\cite{Lin2014}.
For the single-class experiments, we also report the AP for small ($\textrm{AP}_\textrm{S}$) and medium ($\textrm{AP}_\textrm{M}$) detections. 
COCO defines detections as ``small'' if they occupy up to $32 \times 32$ pixels, ``medium'' if they occupy between $32 \times 32$ and $96 \times 96$ pixels, ``large'' otherwise. 
No large detections are present in \ourdataset.
For the multi-class experiments, we report the per-class AP, the average AP (Avg), and the instance-weighted average AP (WAvg).

\subsection{Detectors}

We compare six popular object detectors: 
Faster R-CNN (2015)~\cite{ren2015faster}, 
VarifocalNet (2021)~\cite{zhang2021varifocalnet}, 
GLIP (2022)~\cite{li2022glip}, 
DINO (2023)~\cite{zhang2023dino}, 
ObjectBox (2022)~\cite{zand2022objectbox}, and 
YOLOv8 (2023)~\cite{Jocher2023YOLOv8}.
We use the MMDetection open source library~\cite{mmdetection} for the implementation of the first four methods, whereas we employ the authors' code for ObjectBox and YOLOv8. 
For a fair comparison, we select the algorithm's backbones to have similar number of parameters: we use ResNet50 for Faster R-CNN, VarifocalNet, and DINO; DarkNet for ObjectBox, and YOLOv8, and Swin-T for GLIP. 





\begin{figure*}[t]
\centering
\includegraphics[width=\linewidth]{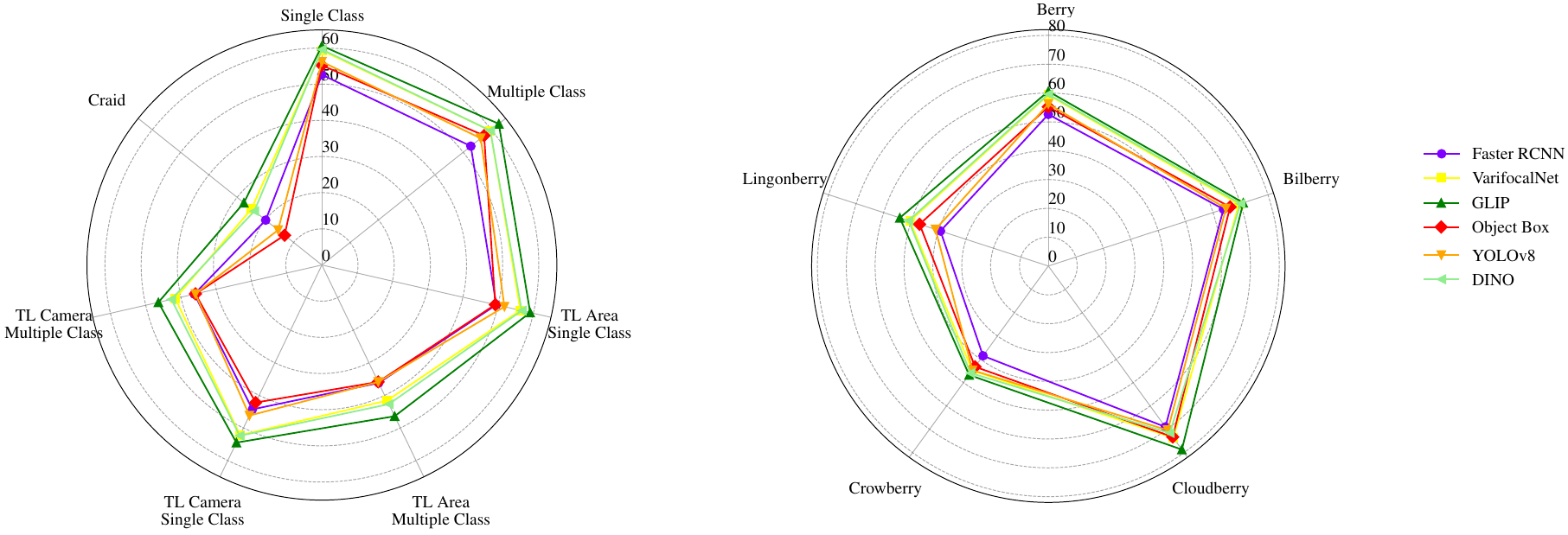}
\caption{Summary of the results quantified in terms of AP. 
(left-hand side) Radar visualisation that includes the evaluation of detectors in coping with single- and multi-class objects, transfer learning (TL) capabilities across different forest areas, TL abilities across different camera sensors, and across different datasets, specifically on CRAID \cite{Akiva2020craid}.
(right-hand side) Radar visualisation that includes the evaluation of detectors for each class of berries.}
\label{fig:radar}
\end{figure*}

\newcolumntype{g}{>{\columncolor{mygray}}c}
\begin{table*}[t] 
\centering
\tabcolsep 4pt
\caption{
Multi-class object detection results evaluated in terms of average precision (AP) and instance-weighted average AP (WAvg).
Key -- Bil.: bilberry, Cloud.: cloudberry, Crow.: crowberry, Lingon.: lingonberry.
Below each berry category, we report the number of instances.
Single-class object detection results evaluated in terms of average precision (AP), AP on small detections ($\textrm{AP}_\textrm{S}$), and AP on medium detections ($\textrm{AP}_\textrm{M}$).
Multi-class to Single-class object detection results (M~2~S) are evaluated in terms of AP.}
\label{tab:sd}
\vspace{-3mm}
\resizebox{\linewidth}{!}{%
\begin{tabular}{rlccccccgggc}
    \toprule
    & \multirow{2}{*}{Algorithm} & \multicolumn{6}{c}{Multi-class} & \multicolumn{3}{c}{\cellcolor{mygray} Single-class} & M 2 S \\ 
    & & Bil. & Cloud. & Crow. & Lingon. & Avg & WAvg & AP & $\textrm{AP}_\textrm{S}$ & $\textrm{AP}_\textrm{M}$ & $\textrm{AP}$ \\
    \midrule
    & \color{gray} Train inst. & \color{gray} 8407 & \color{gray} 683 & \color{gray} 4277 & \color{gray} 3148 & \multicolumn{2}{c}{\color{gray} 16515} & \color{gray} 16515 & \color{gray} 14398 & \color{gray} 2114 & \color{gray} 16515 \\
    & \color{gray} Test inst. & \color{gray} 389 & \color{gray} 42 & \color{gray} 301 & \color{gray} 233 & \multicolumn{2}{c}{\color{gray} 965} & \color{gray} 965 & \color{gray} 875 & \color{gray} 90 & \color{gray} 965 \\
    \small \color{gray} 1 & FasterRCNN & 63.8 & 69.0 & 38.5 & 39.3 & 52.7 & 50.2 & 52.6 & 49.9 & 76.1 & 52.0 \\
    \small \color{gray} 2 & VarifocalNet & 69.4 & 73.7 & 44.8 & 50.4 & 59.6 & 57.3 & 59.3 & 56.9 & 81.8 & 58.4 \\
    \small \color{gray} 3 & GLIP & \textbf{71.0} & \textbf{78.7} & \textbf{46.7} & \textbf{54.1} & \textbf{62.6} & \textbf{59.7} & \textbf{60.6} & \textbf{58.0} & \textbf{82.3} & \textbf{60.9} \\
    \small \color{gray} 4 & DINO & 70.0 & 71.6 & 45.9 & 50.9 & 59.6 & 57.9 & 59.6 & 57.5 & 78.1 & 58.9 \\
    \small \color{gray} 5 & ObjectBox & 66.2 & 73.3 & 43.2 & 46.9 & 57.4 & 54.7 & 55.2 & 52.6 & 77.7 & 56.2 \\
    \small \color{gray} 6 & YOLOv8 & 64.7 & 70.4 & 44.7 & 41.1 & 55.2 & 53.0 & 56.2 & 53.4 & 79.6 & 55.1 \\
    \bottomrule
\end{tabular}
}
\end{table*}

\subsection{Quantitative results}

The qualitative results encompass four experiments. 
In the first experiment, we assess the performance of detectors in coping with single- and multi-class objects.
In the second experiment, we evaluate the detectors' transfer learning capabilities across different forest areas. 
In the third experiment, we examine the detectors' transfer learning abilities across different camera sensors. 
In the fourth experiment, we assess the detectors' transfer learning abilities across different datasets; specifically, we test them on CRAID \cite{Akiva2020craid}. 
Fig.~\ref{fig:radar} summarises the results of these experiments.

\noindent\textbf{Single- and multi-class object detection.}
Tab.~\ref{tab:sd} shows object detection results on \ourdataset's test set.
Rows show the performance of different detectors trained on the training split of \ourdataset.
Columns show different settings: we report single-class, multi-class and multi-class 2 single-class settings.
The first two rows with grey values report the number of train and test bbox instances, that are useful given the imbalance of some classes.
The best performing method is GLIP, closely followed by DINO and VarifocalNet.
In the single-class setting, by comparing $\textrm{AP}_\textrm{S}$ with $\textrm{AP}_\textrm{M}$ we observe medium-sized berries are better detected than small-size ones.
In the multi-class setting, we report results on each class separately in terms of Avg and WAvg in the case of Bilberry (Bil.), Cloudberry (Cloud.), Crowberry (Crow.), and Lingonberry (Lingon.).
The multi-class WAvg is always lower than single-class AP, this is because algorithm localisation is simpler than classification.
The standard deviation for WAvg is 3.5, while for the single-class AP it is 3.1.
For GLIP and ObjectBox the gap between WAvg and single-class AP is small, suggesting that these two methods can classify bounding boxes well.
Cloudberry and bilberry are identified with greater ease compared to crowberry and lingonberry, a disparity stemming from the imbalance in the distribution of training and test data.
Cloudberry accounts for 26\% of the training set but only 4\% of the test set, whereas lingonberry represents 4\% of the training set and 24\% of the test set.
The last column shows that GLIP and ObjectBox benefit from multi-class training, whereas the other methods yield better results when trained on single-class data.

\noindent\textbf{Cross-area transfer learning.}
We evaluate the ability of algorithms to generalise across different areas. 
Training is conducted on three areas, and testing is performed on a fourth, distinct area. 
\cref{tab:tlarea} presents the detection results, with grey columns indicating the number of instances per class. 
Areas 1 and 2 show the lowest scores in terms of Avg. 
However, Area 1 achieves higher scores in terms of WAvg because it has a greater number of bilberry instances, which are easier to classify. 
Low WAvg values for Area 2 can be attributed to the low single-class average precision (AP) value, suggesting that the task of berry localisation does not transfer well to this area. 
Area 4 is found to be the easiest for transfer, primarily due to a higher number of training instances compared to testing instances.
VarifocalNet, GLIP, and DINO consistently perform comparably in terms of Avg, with the exception of Area 2, where the presence of only one test sample of Cloudberry is not considered representative. 
Except for Area 2, bilberry is the easiest class to transfer, indicating that it maintains its visual characteristics across various areas. 
Conversely, Crowberry and Lingonberry are generally the most difficult classes to transfer, suggesting that they may exhibit inconsistent visual properties across geographical areas. 
For example, Areas 1 and 3 contain a larger number of training samples than testing samples for both classes, yet the APs are significantly lower than for the Bilberry class. 
The most challenging classes to classify, namely Crowberry and Lingonberry, also present the greatest challenges in transfer learning, with the exception of Area 4, which features only a few test samples.

\newcolumntype{g}{>{\columncolor{mygray}}r}
\begin{table}[t]
\centering
\tabcolsep 12pt
\caption{
Object detection results in terms of average precision (AP) when performing transfer learning across geographical areas.
Key - B.: bilberry, Cl.: cloudberry, Cr.: crowberry, L.: lingonberry, S.: Single-class.
}

\vspace{-3mm}
\resizebox{\columnwidth}{!}{%
\begin{tabular}{rcclggrrrrrr}
    \toprule
    & \rotatebox{90}{Area} & \rotatebox{90}{Setting} & \rotatebox{90}{AP} & \rotatebox{90}{Train Inst.} & \rotatebox{90}{Test Inst.} & \rotatebox{90}{FasterRCNN} & \rotatebox{90}{VarifocalNet} & \rotatebox{90}{GLIP} & \rotatebox{90}{DINO} & \rotatebox{90}{ObjectBox} & \rotatebox{90}{YOLOv8} \\
    \midrule
    %
    %
    \small \color{gray} 1 & \multirow{7}{*}{\rotatebox{90}{Area 1}} & \multirow{6}{*}{\rotatebox{90}{Multi-class}} & B. & 4357 & 4684 & 63.6 & 69.6 & 69.5 & 68.8 & 64.0 & 63.8 \\
    \small \color{gray} 2 & & & Cl. & 721 & 4 & 0.0 & 0.0 & 0.0 & 0.0 & 0.0 & 0.0 \\
    \small \color{gray} 3 & & & Cr. & 4561 & 17 & 2.8 & 27.5 & 8.5 & 19.2 & 10.6 & 12.4 \\
    \small \color{gray} 4 & & & L. & 3273 & 114 & 13.0 & 13.3 & 15.1 & 16.6 & 11.5 & 9.6 \\
    \small \color{gray} 5 & & & Avg & & & 19.9 & \textbf{27.6} & 23.3 & 26.2 & 21.5 & 21.5 \\
    \small \color{gray} 6 & & & WAvg & \raisebox{.5\normalbaselineskip}[0pt][0pt]{12912} & \raisebox{.5\normalbaselineskip}[0pt][0pt]{4819} & 62.1 & \textbf{68.1} & 67.9 & 67.3 & 62.5 & 62.3 \\
    \cmidrule{3-12}
    \small \color{gray} 7 & & \rotatebox{90}{S.} & AP & 12912 & 4819 & 63.6 & 68.3 & \textbf{69.0} & 68.6 & 62.9 & 64.6 \\
    \midrule
    %
    %
    \small \color{gray} 8 & \multirow{7}{*}{\rotatebox{90}{Area 2}} & \multirow{6}{*}{\rotatebox{90}{Multi-class}} & B. & 8619 & 188 & 26.2 & 31.3 & 32.1 & 32.0 & 28.3 & 29.5 \\
    \small \color{gray} 9 & & & Cl. & 724 & 1 & 3.6 & 80.0 & 90.0 & 0.0 & 0.0 & 0.5 \\
    \small \color{gray} 10 & & & Cr. & 3178 & 1470 & 16.3 & 18.6 & 23.6 & 20.9 & 17.7 & 19.3 \\
    \small \color{gray} 11 & & & L. & 1956 & 1515 & 33.3 & 41.2 & 44.7 & 41.6 & 31.3 & 32.4\\
    \small \color{gray} 12 & & & Avg & & & 19.9 & 42.8 & \textbf{47.6} & 23.6 & 19.3 & 20.4 \\
    \small \color{gray} 13 & & & WAvg & \raisebox{.5\normalbaselineskip}[0pt][0pt]{14477} & \raisebox{.5\normalbaselineskip}[0pt][0pt]{3174} & 25.0 & 30.2 & \textbf{34.2} & 31.4 & 24.8 & 26.2 \\
    \cmidrule{3-12}
    \small \color{gray} 14 & & \rotatebox{90}{S.} & AP & 14477 & 3174 & 22.9 & 28.8 & \textbf{33.6} & 30.2 & 22.4 & 26.3 \\
    \midrule
    %
    %
    \small \color{gray} 15 & \multirow{7}{*}{\rotatebox{90}{Area 3}} & \multirow{6}{*}{\rotatebox{90}{Multi-class}} & B. & 8339 & 472 & 52.4 & 60.7 & 64.2 & 63.8 & 57.6 & 60.9 \\
    \small \color{gray} 16 & & & Cl. & 130 & 628 & 64.2 & 73.2 & 76.5 & 72.9 & 64.7 & 53.3 \\
    \small \color{gray} 17 & & & Cr. & 3772 & 857 & 34.1 & 45.6 & 49.6 & 49.7 & 40.7 & 45.3 \\
    \small \color{gray} 18 & & & L. & 2849 & 545 & 13.2 & 33.2 & 31.0 & 33.9 & 12.9 & 15.0 \\
    \small \color{gray} 19 & & & Avg & & & 41.0 & 53.2 & \textbf{55.3} & 55.1 & 44.0 & 43.6 \\
    \small \color{gray} 20 & & & WAvg & \raisebox{.5\normalbaselineskip}[0pt][0pt]{15090} & \raisebox{.5\normalbaselineskip}[0pt][0pt]{2502} & 40.6 & 52.7 & \textbf{55.1} & 54.7 & 43.9 & 43.7 \\
    \cmidrule{3-12}
    \small \color{gray} 21 & & \rotatebox{90}{S.} & AP & 15090 & 2502 & 44.5 & 56.0 & \textbf{60.9} & 56.9 & 45.8 & 48.8 \\
    \midrule
    %
    %
    \small \color{gray} 22 & \multirow{7}{*}{\rotatebox{90}{Area 4}} & \multirow{6}{*}{\rotatebox{90}{Multi-class}} & B. & 8310 & 506 & 68.2 & 72.1 & 73.9 & 71.4 & 68.0 & 67.8 \\
    \small \color{gray} 23 & & & Cl. & 725 & 0 & - & - & - & - & - & - \\
    \small \color{gray} 24 & & & Cr. & 4578	& 0 & - & - & - & - & - & - \\
    \small \color{gray} 25 & & & L. & 3329 & 59 & 57.4 & 62.4 & 58.1 & 57.9 & 49.0 & 54.2 \\
    \small \color{gray} 26 & & & Avg & & & 62.8 & \textbf{67.3} & 66.0 & 64.7 & 58.5 & 61.0 \\
    \small \color{gray} 27 & & & WAvg & \raisebox{.5\normalbaselineskip}[0pt][0pt]{16942} & \raisebox{.5\normalbaselineskip}[0pt][0pt]{565} & 67.1 & 71.1 & \textbf{72.3} & 70.0 & 66.0 & 66.4 \\
    \cmidrule{3-12}
    \small \color{gray} 28 & & \rotatebox{90}{S.} & AP & 16942 & 565 & 67.0 & 71.9 & \textbf{72.8} & 71.1 & 65.7 & 67.4 \\
    \bottomrule
\end{tabular}
}
\label{tab:tlarea}
\end{table}

\noindent\textbf{Cross-camera transfer learning.}
We evaluate the ability of algorithms to transfer across different cameras. \cref{tab:tlcamera} displays the detection results when algorithms are trained with data captured by four cameras and then tested on a different camera (\cref{tab:sensors}). 
In the multi-class scenario, GLIP outperforms other algorithms in terms of Avg, followed by DINO and VerifocalNet. 
Faster-RCNN reports the lowest average in most cases. 
In the single-class scenario, GLIP again outperforms in transfer learning cases, except in the case of the DJI Mini 2, where DINO performs best. 
When the number of test instances is imbalanced (\eg, in the case of DJI Mavic 2 Pro), APs vary significantly, yet the average APs does not accurately reflect each algorithm's performance. 
In such cases, WAvg is a more rational metric to prevent class bias. 
Comparing WAvg amongst the four classes with AP of a single class, we observe three behaviours. 
Firstly, in the case of a single class, AP remains higher than the multiclass WAvg (\eg, Xiaomi 12X), as the single-class AP reflects only a localisation task, while WAvg scores also include an additional classification task that may hinder performance. 
Secondly, it is often observed that the single-class APs and the multiclass WAvg are comparable (\eg., DJI Mavic 2 Pro), indicating that the algorithms perform well both in classification and localisation. 
Thirdly, a higher WAvg than single-class AP suggests that classification can enhance the algorithm's learning in localisation abilities (\eg, GoPro 11).
Overall, considering both the single class APs as well as the multi-class WAvg, we observe that DJI Mavic 2 Pro is the easiest sensor to transfer to, probably since its test set is dominated by samples from the Bilberry class that report the highest score amongst the four classes.
Oppositely, DJI Mavic 3M is the hardest sensor to transfer to as it contains several Crowberry and Lingonberry instances that are the most difficult to handle.

\newcolumntype{g}{>{\columncolor{mygray}}r}
\begin{table}[t!]
\centering
\tabcolsep 12pt
\caption{
Detection results when performing transfer learning across different sensors.
Key - B.: bilberry, Cl.: cloudberry, Cr.: crowberry, L.: lingonberry, S.: Single-class.
}
\label{tab:tlsensors}
\vspace{-3mm}
\resizebox{\columnwidth}{!}{
\begin{tabular}{rcclggrrrrrr}
    \toprule
    & \rotatebox{90}{Camera} & \rotatebox{90}{Setting} & \rotatebox{90}{AP} & \rotatebox{90}{Train Inst.} & \rotatebox{90}{Test Inst.} & \rotatebox{90}{FasterRCNN} & \rotatebox{90}{VarifocalNet} & \rotatebox{90}{GLIP} & \rotatebox{90}{ObjectBox} & \rotatebox{90}{YOLOv8} & \rotatebox{90}{DINO} \\
    \midrule

    \small \color{gray} 1 & \multirow{7}{*}{\rotatebox{90}{DJI Mavic 2 Pro}} & \multirow{6}{*}{\rotatebox{90}{Multi-class}} & B. & 3871 & 5190 & 61.7 & 66.0 & 67.9 & 60.9 & 62.0 & 66.1\\
    \small \color{gray} 2 & & & Cl. & 721 & 4 & 0.0 & 0.0 & 0.0 & 0.0 & 0.0 & 0.0 \\
    \small \color{gray} 3 & & & Cr. & 4561	& 17 & 2.5 & 14.9 & 20.1 & 12.30 & 15.4 & 22.0 \\
    \small \color{gray} 4 & & & L. & 3221 & 173 & 27.8 & 29.9 & 32.6 & 24.8 & 24.3 & 32.9 \\
    \small \color{gray} 5 & & & Avg &  &  & 23.0 & 27.9 & \textbf{30.6} & 24.5 & 25.4 & 30.3 \\

    \small \color{gray} 6 & & & WAvg & \raisebox{.5\normalbaselineskip}[0pt][0pt]{12374} & \raisebox{.5\normalbaselineskip}[0pt][0pt]{5384} & 60.3 &	65.4 &	\textbf{66.5} &	59.5 &	60.6 &	64.8 \\
    
    \cmidrule{3-12}
    \small \color{gray} 7 & & \rotatebox{90}{S.} & AP & & &  61.2 & 67.2 & \textbf{67.8} & 59.4 & 61.6 & 66.5 \\    
    \midrule

    \small \color{gray} 8 & \multirow{7}{*}{\rotatebox{90}{DJI Mavic 3M}} & \multirow{6}{*}{\rotatebox{90}{Multi-class}} & B. & 8162 & 660 & 45.3 & 53.4 & 56.1 & 49.1 & 51.5 & 53.2 \\
    \small \color{gray} 9 & & & Cl. & 129 & 629 & 63.5 & 73.3 & 77.6 & 62.6 & 58.4 & 68.9 \\
    \small \color{gray} 10 & & & Cr. & 2372	& 2327 & 21.5 & 26.0 & 30.7 & 22.3 & 27.2 & 29.2 \\
    \small \color{gray} 11 & & & L. & 1424 & 2060 & 26.4 & 30.7 & 43.2 & 25.9 & 24.1 & 34.6 \\
    \small \color{gray} 12 & & & Avg & & & 39.2 & 45.8 & \textbf{51.9} & 40.0 & 40.3 & 46.5 \\
    \small \color{gray} 13 & & & WAvg & \raisebox{.5\normalbaselineskip}[0pt][0pt]{12087} & \raisebox{.5\normalbaselineskip}[0pt][0pt]{5676} & 30.7 &	36.1 &	\textbf{43.3} &	31.1 &	32.3 &	38.3 \\
    
    \cmidrule{3-12}
    \small \color{gray} 14 & & \rotatebox{90}{S.} & AP  & & &  31.9 & 41.0 & \textbf{44.8} & 33.50 & 35.6 & 41.4 \\    
    \midrule

    \small \color{gray} 15 & \multirow{7}{*}{\rotatebox{90}{DJI Mini 2}} & \multirow{6}{*}{\rotatebox{90}{Multi-class}} & B. & 8768 & 28 & 30.9 & 35.5 & 43.8 & 40.0 & 35.5 & 33.9 \\
    \small \color{gray} 16 & & & Cl. & 646 & 82 & 72.8 & 73.5 & 71.5 & 71.2 & 74.0 & 75.1 \\
    \small \color{gray} 17 & & & Cr. & 4515	& 66 & 30.0 & 35.5 & 32.0 & 32.6 & 31.8 & 36.5 \\
    \small \color{gray} 18 & & & L. & 3381 & 0.0 & - & - &- & - & - & - \\
    \small \color{gray} 19 & & & Avg & & & 44.6 & 48.2 & \textbf{49.1} & 47.9 & 47.1 & 48.5 \\
    \small \color{gray} 20 & & & WAvg & \raisebox{.5\normalbaselineskip}[0pt][0pt]{17310} & \raisebox{.5\normalbaselineskip}[0pt][0pt]{176} &  50.0 &	53.2 &	52.2 &	51.7 &	52.0 &	\textbf{54.0} \\
    
    \cmidrule{3-12}
    \small \color{gray} 21 & & \rotatebox{90}{S.} & AP & & &  49.7 & 55.1 & 51.6 & 27.4 & 50.5 &  \textbf{56.3} \\    
    \midrule

    \small \color{gray} 22 & \multirow{7}{*}{\rotatebox{90}{GoPro 11}} & \multirow{6}{*}{\rotatebox{90}{Multi-class}} & B. & 8794 & 2.0 & 35.5 & 44.6 & 63.5 & 10.1 & 40.4 & 45.4 \\   
    \small \color{gray} 23 & & & Cl. & 693 & 34 & 51.2 & 59.0 & 57.4 & 51.5 & 42.1 & 52.4 \\
    \small \color{gray} 24 & & & Cr. & 4575 & 3.0 & 20.6 & 10.4 & 43.7 & 0.4 & 13.7 & 29.2 \\   
    \small \color{gray} 25 & & & L. & 3378 & 3.0 & 36.2 & 49.2 & 43.1 & 29.1 & 6.5 & 46.7 \\
    \small \color{gray} 26 & & & Avg & & & 35.9 & 40.8 & \textbf{51.9} & 22.8 & 25.7 & 43.4 \\
    \small \color{gray} 27 & & & WAvg & \raisebox{.5\normalbaselineskip}[0pt][0pt]{17440} & \raisebox{.5\normalbaselineskip}[0pt][0pt]{42} &  47.2 &	54.1 &	\textbf{55.6} &	44.2 &	37.4 &	50.0 \\
    
    \cmidrule{3-12}
    \small \color{gray} 28 & & \rotatebox{90}{S.} & AP & & &  32.4 & 45.0 & \textbf{52.4} & 44.2 & 37.3 & 44.6 \\
    \midrule

    \small \color{gray} 29 & \multirow{7}{*}{\rotatebox{90}{Xiaomi 12X}} & \multirow{6}{*}{\rotatebox{90}{Multi-class}} & B. & 5589 & 3350 & 47.4 & 53.7 & 55.8 & 51.0 & 50.0 & 55.4 \\
    \small \color{gray} 30 & & & Cl. & 711 & 15 & 43.0 & 55.6 & 49.1 & 46.30 & 44.7 & 43.6 \\
    \small \color{gray} 31 & & & Cr. & 2289 & 2358 & 23.8 & 28.1 & 40.0 & 37.5 & 28.6 & 32.1 \\
    \small \color{gray} 32 & & & L. & 2120 & 1335 & 38.5 & 44.5 & 49.7 & 41.9 & 39.8 & 48.0 \\
    \small \color{gray} 33 & & & Avg & & & 38.2 & 45.5 & \textbf{48.6} & 44.2 & 40.8 & 44.7 \\
    \small \color{gray} 34 & & & WAvg & \raisebox{.5\normalbaselineskip}[0pt][0pt]{10709} & \raisebox{.5\normalbaselineskip}[0pt][0pt]{7058} &  37.8 &	43.4 &	\textbf{49.3} &	44.7 &	40.9 &	46.1 \\
    
    \cmidrule{3-12}
    \small \color{gray} 35 & & \rotatebox{90}{S.} & AP  & & &  46.2 & 52.7 & \textbf{56.0} & 46.30 & 46.2 & 53.1 \\
    
    \bottomrule
\end{tabular}
}
\label{tab:tlcamera}
\end{table}

\begin{table}[b!]
\centering
\tabcolsep 5pt
\caption{
Object detection results in terms of average precision (AP) when training on \ourdataset and testing on CRAID~\cite{Akiva2020craid}.
}
\label{tab:f2c}

\vspace{-3mm}
\begin{tabular}{cccccc}
    \toprule
    FasterRCNN & VarifocalNet & GLIP & DINO & ObjectBox & YOLOv8 \\
    \midrule
    19.9 & 24.9 & \textbf{27.6} & 13.2 & 15.5 & 23.9 \\
    \bottomrule
\end{tabular}
\end{table}

\begin{figure*}[t]
\centering

    \begin{minipage}{0.235\textwidth}
        \centering
        \footnotesize VarifocalNet~\cite{zhang2021varifocalnet}
    \end{minipage}
    \begin{minipage}{0.235\textwidth}
        \centering
        \footnotesize DINO~\cite{zhang2023dino}
    \end{minipage}
    \begin{minipage}{0.235\textwidth}
        \centering
        \footnotesize GLIP~\cite{li2022glip}
    \end{minipage}
    \begin{minipage}{0.235\textwidth}
        \centering
        \footnotesize Ground truth
    \end{minipage}

    %
    %
    \smallskip
    \begin{minipage}{0.235\textwidth}
        \centering
        \begin{overpic}[width=\textwidth]{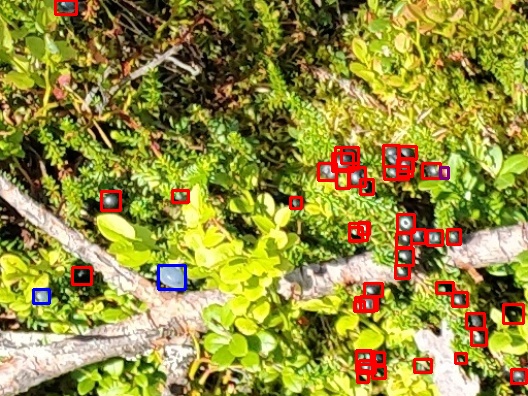}
        \end{overpic}
    \end{minipage}
    \begin{minipage}{0.235\textwidth}
        \centering
        \begin{overpic}[width=\textwidth]{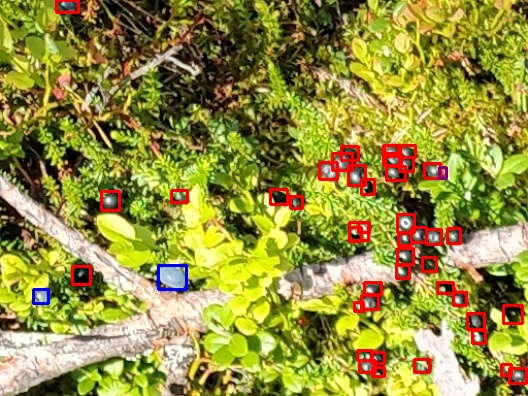}
        \end{overpic}
    \end{minipage}
    \begin{minipage}{0.235\textwidth}
        \centering
        \begin{overpic}[width=\textwidth]{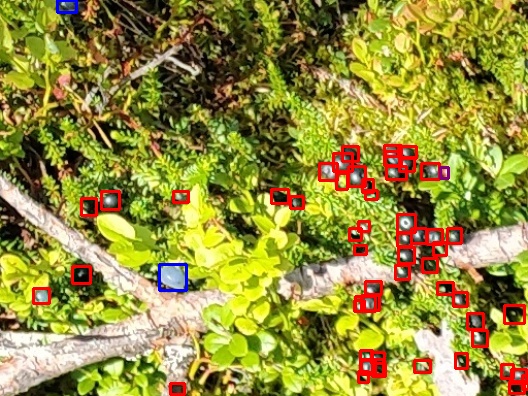}
        \end{overpic}
    \end{minipage}
    \begin{minipage}{0.235\textwidth}
        \centering
        \begin{overpic}[width=\textwidth]{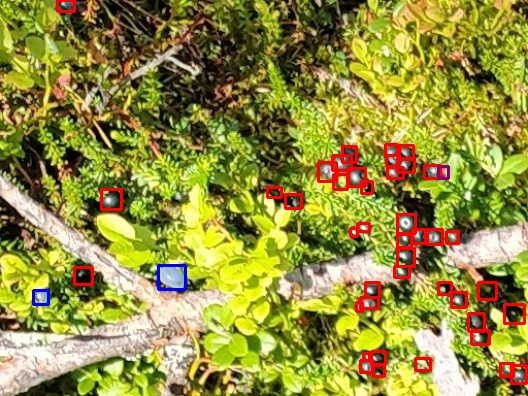}
        \end{overpic}
    \end{minipage}

    %
    %
    \smallskip
    \begin{minipage}{0.235\textwidth}
        \centering
        \begin{overpic}[width=\textwidth]{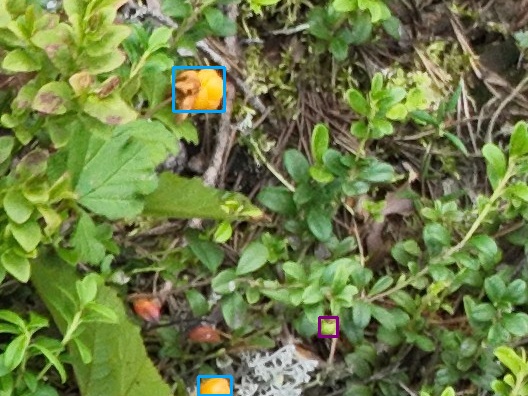}
        \end{overpic}
    \end{minipage}
    \begin{minipage}{0.235\textwidth}
        \centering
        \begin{overpic}[width=\textwidth]{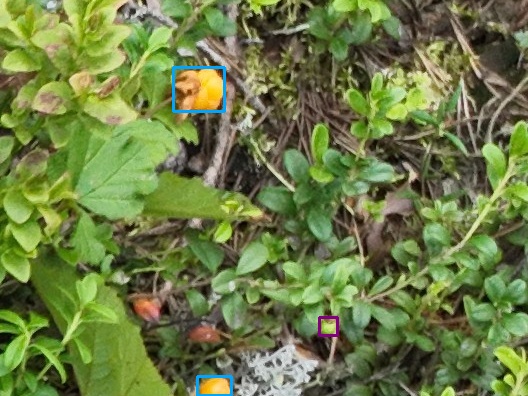}
        \end{overpic}
    \end{minipage}
    \begin{minipage}{0.235\textwidth}
        \centering
        \begin{overpic}[width=\textwidth]{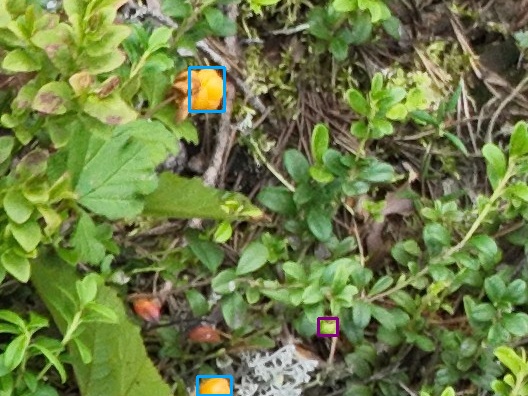}
        \end{overpic}
    \end{minipage}
    \begin{minipage}{0.235\textwidth}
        \centering
        \begin{overpic}[width=\textwidth]{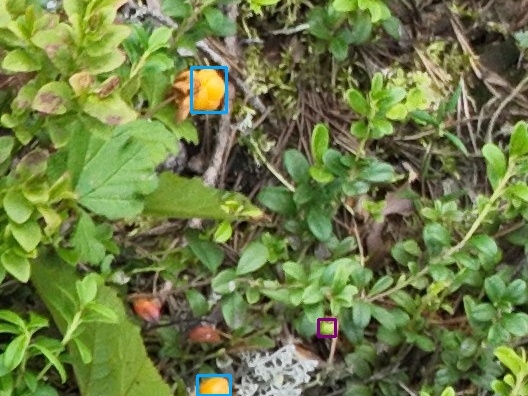}
        \end{overpic}
    \end{minipage}

    %
    %
    \smallskip
    \begin{minipage}{0.235\textwidth}
        \centering
        \begin{overpic}[width=\textwidth]{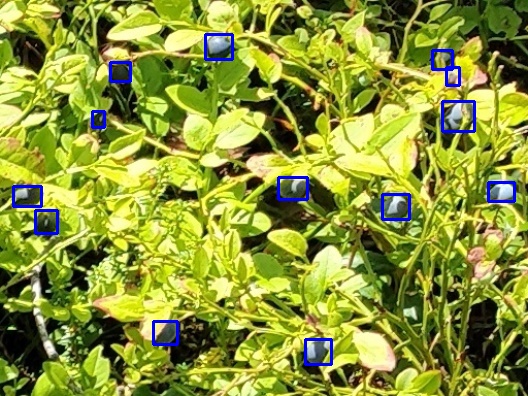}
        \end{overpic}
    \end{minipage}
    \begin{minipage}{0.235\textwidth}
        \centering
        \begin{overpic}[width=\textwidth]{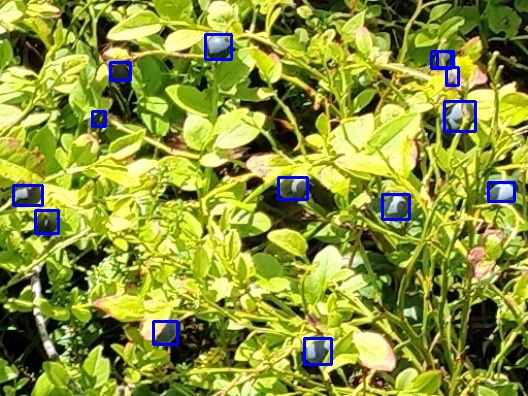}
        \end{overpic}
    \end{minipage}
    \begin{minipage}{0.235\textwidth}
        \centering
        \begin{overpic}[width=\textwidth]{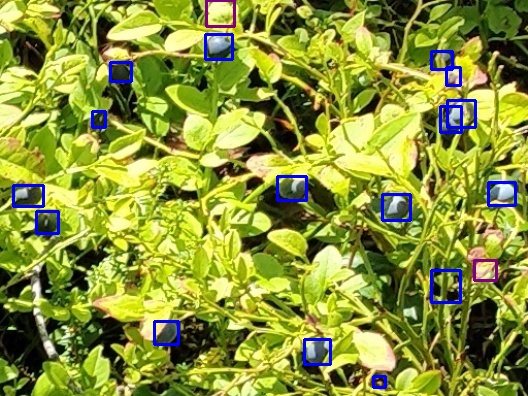}
        \end{overpic}
    \end{minipage}
    \begin{minipage}{0.235\textwidth}
        \centering
        \begin{overpic}[width=\textwidth]{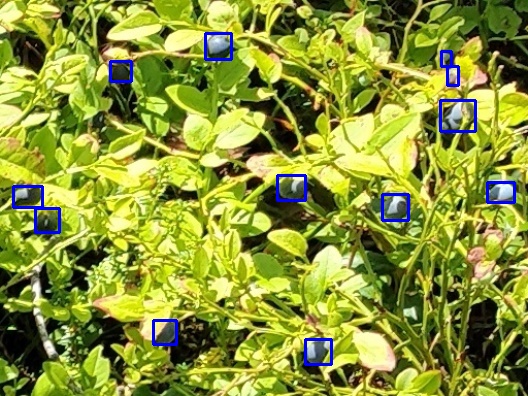}
        \end{overpic}
    \end{minipage}

    %
    %
    \smallskip
    \begin{minipage}{0.235\textwidth}
        \centering
        \begin{overpic}[width=\textwidth]{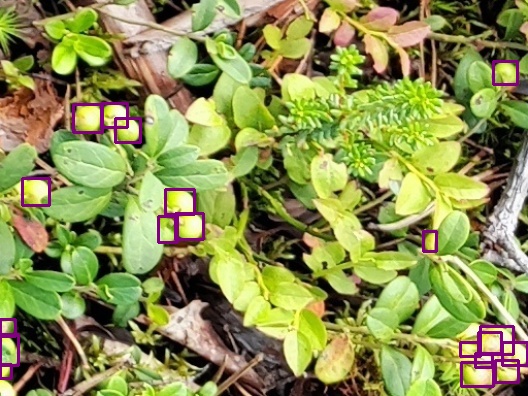}
        \end{overpic}
    \end{minipage}
    \begin{minipage}{0.235\textwidth}
        \centering
        \begin{overpic}[width=\textwidth]{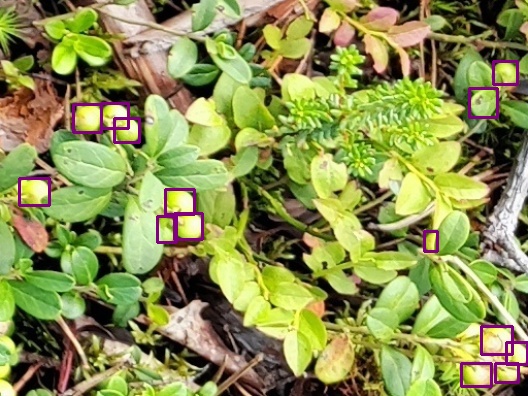}
        \end{overpic}
    \end{minipage}
    \begin{minipage}{0.235\textwidth}
        \centering
        \begin{overpic}[width=\textwidth]{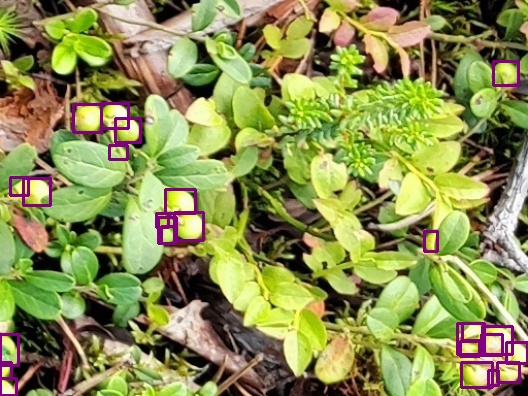}
        \end{overpic}
    \end{minipage}
    \begin{minipage}{0.235\textwidth}
        \centering
        \begin{overpic}[width=\textwidth]{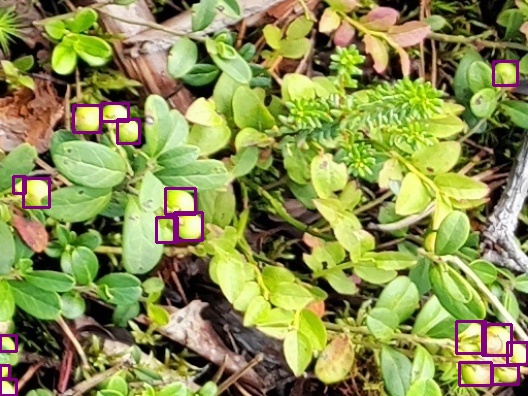}
        \end{overpic}
    \end{minipage}

    \vspace{-2mm}
    \caption{
    Qualitative results.
    Columns show a comparison against the best-performing methods (first three columns) and the ground-truth reference (last column).
    Rows show different examples.
    Key -- Red: bilberry, azure: cloudberry, blue: crowberry, purple: lingonberry.
    }
    \label{fig:qual_sd}
\end{figure*}

\noindent\textbf{Cross-dataset transfer learning.}
We evaluate the algorithms' ability to generalise across datasets, \ie, trained on \ourdataset and tested on the 702 images of the test set of CRAID \cite{Akiva2020craid}. 
CRAID contains only cranberries, which are not included in \ourdataset. 
Moreover, CRAID is captured with a different sensor, in various geographical locations, and under different acquisition conditions. 
\cref{tab:f2c} presents the results. 
Consistent with previous results, GLIP performs the best. 
In contrast, DINO, which ranks second-best in prior evaluations, exhibits a significant performance drop on CRAID. 
Overall, this cross-dataset experiment shows a notable decline in performance for all the algorithms, attributable to the severe domain discrepancy between the training set (\ourdataset) and the test set.

\begin{figure*}[ht]
\centering

    \hspace{2mm}
    \begin{minipage}{0.235\textwidth}
        \centering
        \footnotesize VarifocalNet~\cite{zhang2021varifocalnet}
    \end{minipage}
    \begin{minipage}{0.235\textwidth}
        \centering
        \footnotesize DINO~\cite{zhang2023dino}
    \end{minipage}
    \begin{minipage}{0.235\textwidth}
        \centering
        \footnotesize GLIP~\cite{li2022glip}
    \end{minipage}
    \begin{minipage}{0.235\textwidth}
        \centering
        \footnotesize Ground truth
    \end{minipage}

    %
    %
    \smallskip
    \hspace{2mm}
    \begin{minipage}{0.235\textwidth}
        \centering
        \begin{overpic}[width=\textwidth]{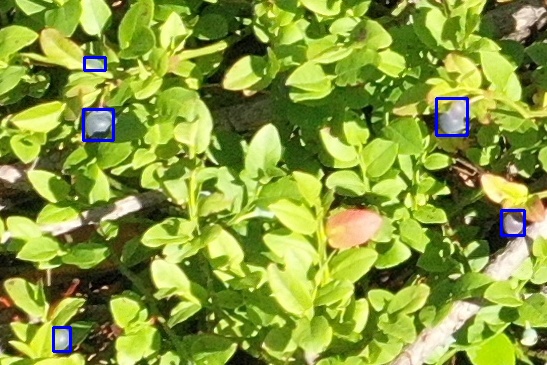}
            \put(-10,20){\rotatebox{90}{\footnotesize Area 1}}
        \end{overpic}
    \end{minipage}
    \begin{minipage}{0.235\textwidth}
        \centering
        \begin{overpic}[width=\textwidth]{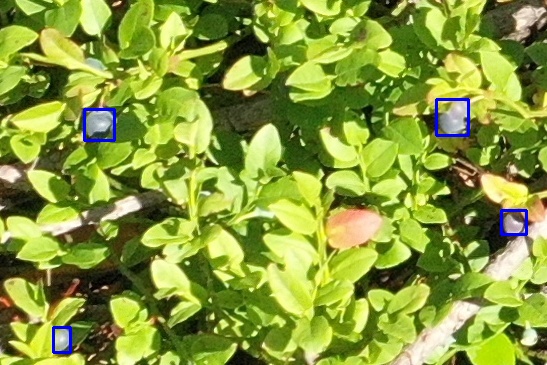}
        \end{overpic}
    \end{minipage}
    \begin{minipage}{0.235\textwidth}
        \centering
        \begin{overpic}[width=\textwidth]{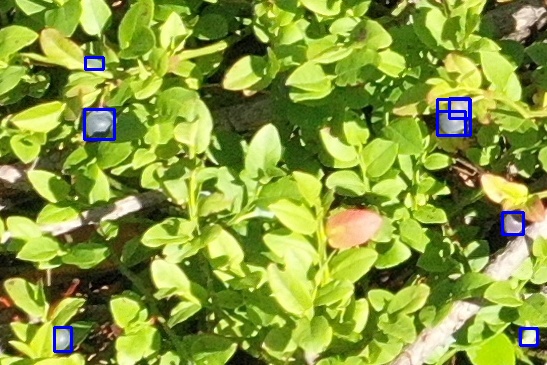}
        \end{overpic}
    \end{minipage}
    \begin{minipage}{0.235\textwidth}
        \centering
        \begin{overpic}[width=\textwidth]{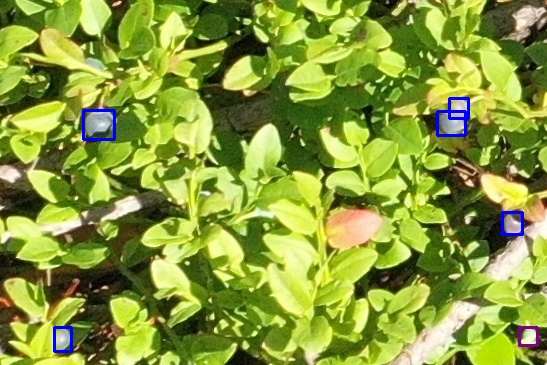}
        \end{overpic}
    \end{minipage}

    %
    %
    \smallskip
    \hspace{2mm}
    \begin{minipage}{0.235\textwidth}
        \centering
        \begin{overpic}[width=\textwidth]{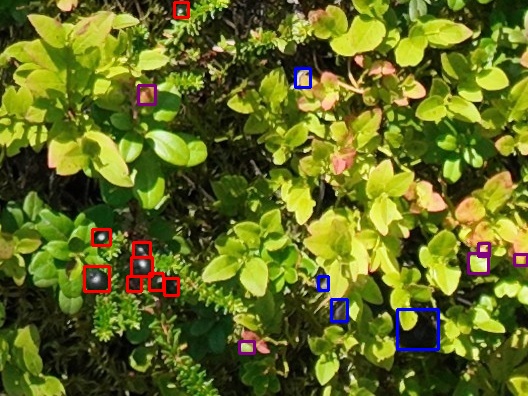}
            \put(-10,25){\rotatebox{90}{\footnotesize Area 2}}
        \end{overpic}
    \end{minipage}
    \begin{minipage}{0.235\textwidth}
        \centering
        \begin{overpic}[width=\textwidth]{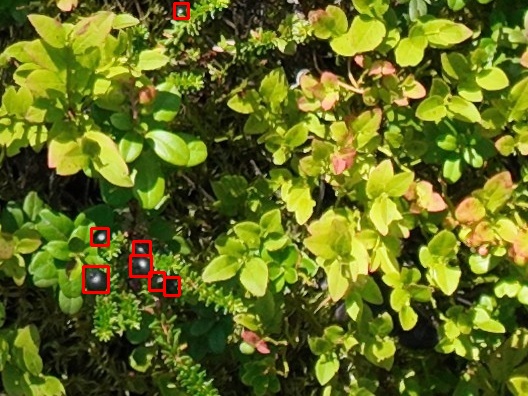}
        \end{overpic}
    \end{minipage}
    \begin{minipage}{0.235\textwidth}
        \centering
        \begin{overpic}[width=\textwidth]{images/qual/tlarea/area2/dino_DJI_20230727103838_0176_D_id56_w3168-3696_h1980-2376}
        \end{overpic}
    \end{minipage}
    \begin{minipage}{0.235\textwidth}
        \centering
        \begin{overpic}[width=\textwidth]{images/qual/tlarea/area2/dino_DJI_20230727103838_0176_D_id56_w3168-3696_h1980-2376}
        \end{overpic}
    \end{minipage}

    \vspace{-2mm}
    \caption{
    Qualitative results for the cross-area transfer learning experiment.
    Columns: a comparison against the best-performing methods (first three columns) and the ground-truth reference (last column).
    Rows: different geographical areas: the first row contains an image captured in Area 1, while the second row shows an image collected in Area 2.
    Key -- Red: bilberry, blue: crowberry, purple: lingonberry.
    }
    \label{fig:qual_tlarea}
\end{figure*}

\begin{figure*}[!ht]
\centering

    \hspace{2mm}
    \begin{minipage}{0.235\textwidth}
        \centering
        \footnotesize VarifocalNet~\cite{zhang2021varifocalnet}
    \end{minipage}
    \begin{minipage}{0.235\textwidth}
        \centering
        \footnotesize DINO~\cite{zhang2023dino}
    \end{minipage}
    \begin{minipage}{0.235\textwidth}
        \centering
        \footnotesize GLIP~\cite{li2022glip}
    \end{minipage}
    \begin{minipage}{0.235\textwidth}
        \centering
        \footnotesize Ground truth
    \end{minipage}

    %
    %
    \smallskip
    \hspace{2mm}
    \begin{minipage}{0.235\textwidth}
        \centering
        \begin{overpic}[width=\textwidth]{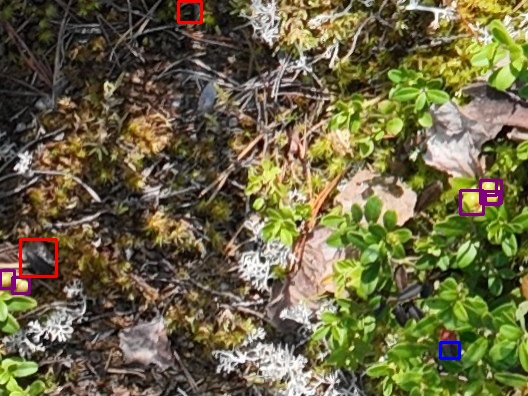}
            \put(-10,2){\rotatebox{90}{\footnotesize DJI Mavic 3M}}
        \end{overpic}
    \end{minipage}
    \begin{minipage}{0.235\textwidth}
        \centering
        \begin{overpic}[width=\textwidth]{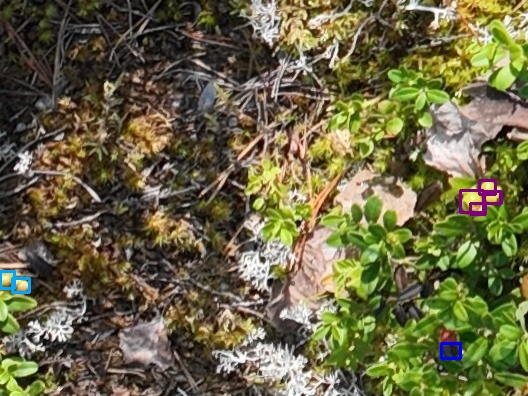}
        \end{overpic}
    \end{minipage}
    \begin{minipage}{0.235\textwidth}
        \centering
        \begin{overpic}[width=\textwidth]{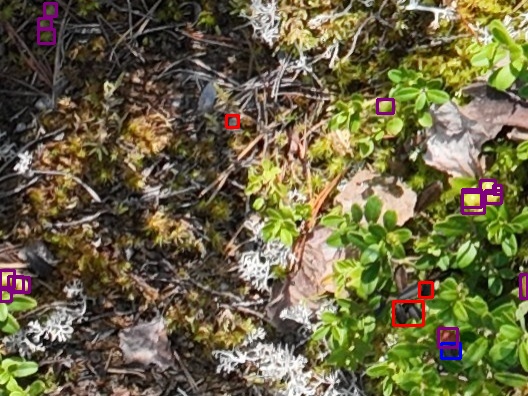}
        \end{overpic}
    \end{minipage}
    \begin{minipage}{0.235\textwidth}
        \centering
        \begin{overpic}[width=\textwidth]{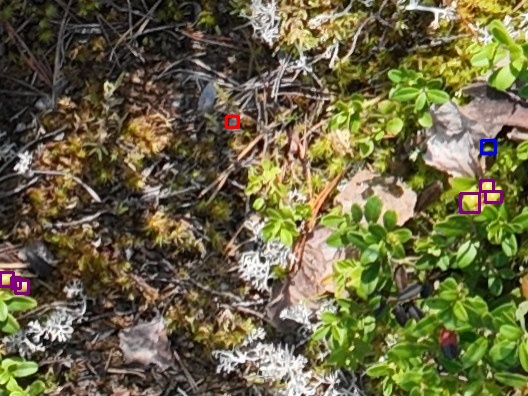}
        \end{overpic}
    \end{minipage}

    %
    %
    \smallskip
    \hspace{2mm}
    \begin{minipage}{0.235\textwidth}
        \centering
        \begin{overpic}[width=\textwidth]{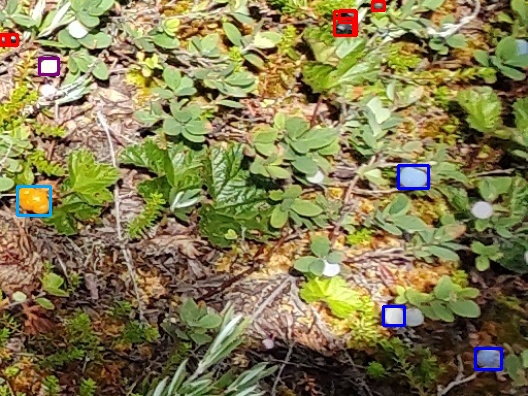}
            \put(-10,7){\rotatebox{90}{\footnotesize Xiaomi 12X}}
        \end{overpic}
    \end{minipage}
    \begin{minipage}{0.235\textwidth}
        \centering
        \begin{overpic}[width=\textwidth]{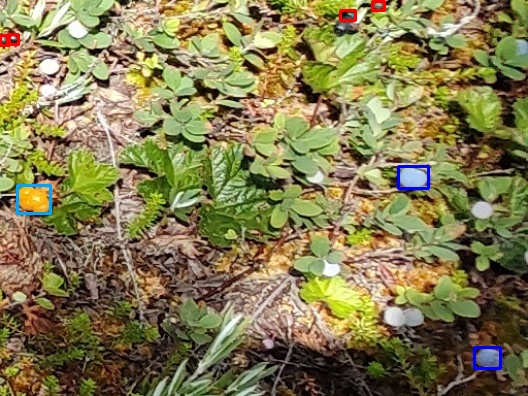}
        \end{overpic}
    \end{minipage}
    \begin{minipage}{0.235\textwidth}
        \centering
        \begin{overpic}[width=\textwidth]{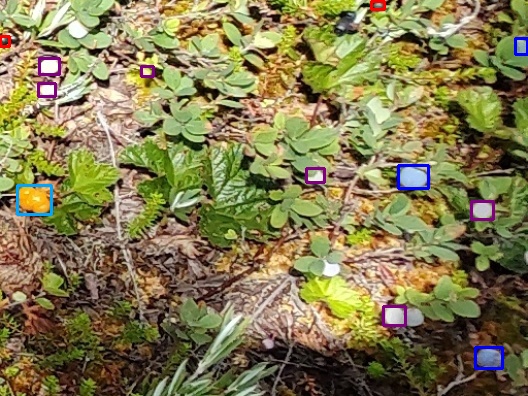}
        \end{overpic}
    \end{minipage}
    \begin{minipage}{0.235\textwidth}
        \centering
        \begin{overpic}[width=\textwidth]{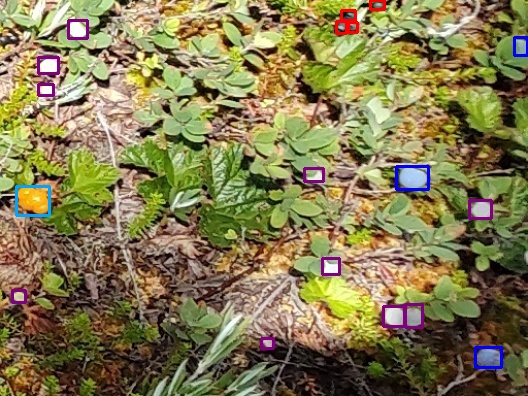}
        \end{overpic}
    \end{minipage}

    \vspace{-2mm}
    \caption{
    Qualitative results for the cross-sensor transfer learning experiment.
    Columns: comparison against the best-performing methods (first three columns) and the ground-truth reference (last column).
    Rows: different sensor types: a DJI Mavic 3M camera for the first row, a Xiaomi 12X smartphone for the second row.
    Key -- Red: bilberry, azure: cloudberry, blue: crowberry, purple: lingonberry.
    }
    \label{fig:qual_tsensor}
\end{figure*}

\begin{figure*}[ht]
\centering

  \begin{tabular}{@{}c@{\,\,}c@{\,\,}c@{\,\,}c}
    \begin{overpic}[width=.245\linewidth]{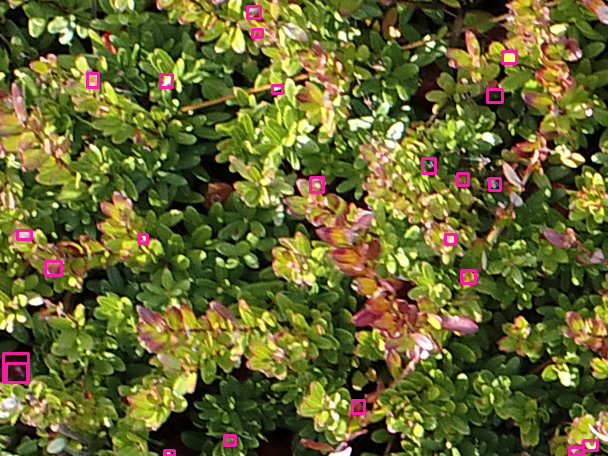}
    \put(32,77){{\small GLIP~\cite{li2022glip}}}
    \end{overpic} &
    \begin{overpic}[width=.245\linewidth]{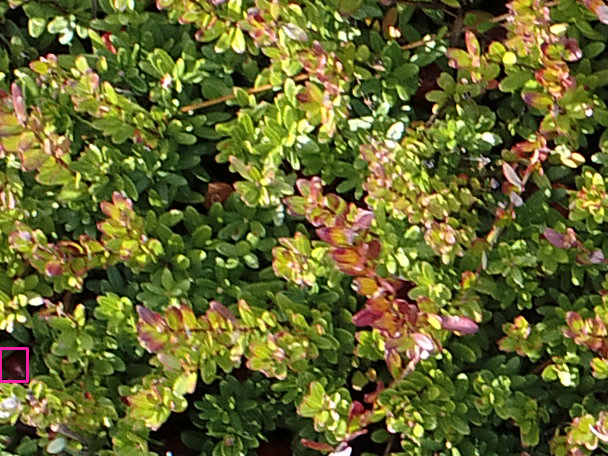}
    \put(18,77){{\small Ground truth}}
    \end{overpic} &
    \begin{overpic}[width=.245\linewidth]{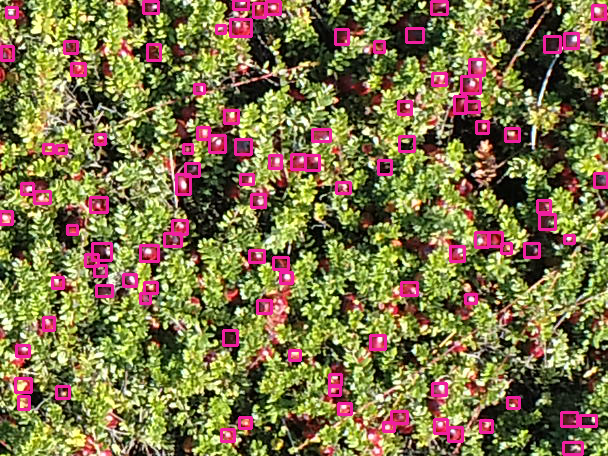}
    \put(32,77){{\small GLIP~\cite{li2022glip}}}
    \end{overpic} &
    \begin{overpic}[width=.245\linewidth]{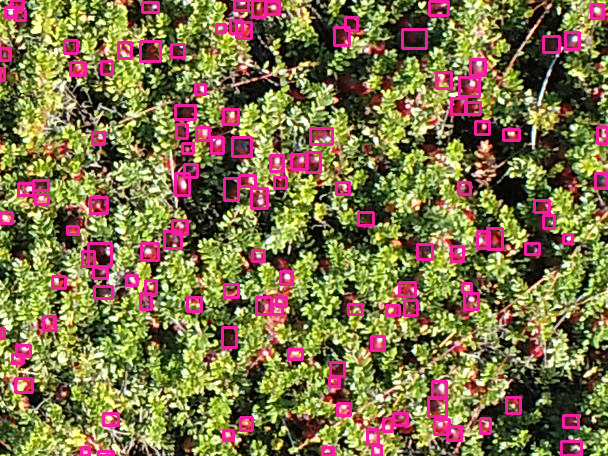}
    \put(18,77){{\small Ground truth}}
    \end{overpic} 
  \end{tabular}
    \vspace{-3mm}
    \caption{
    Qualitative results for the cross-datasets transfer learning experiment.
    We compare GLIP predictions (first column) to ground-truth detections (second column).
    First columns contain a single isolated berry. Second two columns show a bush with several berries.}
\label{fig:qual_craid}
\end{figure*}

\subsection{Qualitative results}
\cref{fig:qual_sd} presents qualitative results for GLIP, DINO, and VarifocalNet, the three best-performing algorithms. 
There are false positive detections within the Bilberry class. 
GLIP misclassifies two instances of Bilberry as Crowberry (in the first row, both at the top and bottom left of the image). 
Then, GLIP identifies two false positives for Bilberry in the third row.
\cref{fig:qual_tlarea} illustrates examples of cross-area transfer learning. 
GLIP accurately detects all true positives but also generates a false positive. VarifocalNet and DINO, on the other hand, miss some berries but yield fewer false positives (see first row).
\cref{fig:qual_tsensor} showcases examples from cross-sensor scenarios. 
The first row highlights that GLIP generates more false positives, and DINO misclassifies two samples.
In the second row, both GLIP and DINO miss some Bilberry instances. 
However, GLIP demonstrates superior performance in handling Lingonberry instances compared to the other two algorithms.
In Fig.~\ref{fig:qual_craid}, we report two detection examples of GLIP when tested on CRAID.
Despite GLIP demonstrating good transfer capabilities in the image of the second row, in the first row we can observe a high number of false positive detections, which are likely the cause of the poor performance reported in Tab~\ref{tab:f2c}.

\section{Conclusions}

We presented \ourdataset, a new image dataset of wild berries collected in Finnish forests using drone technology, marking a significant advancement for the automation of berry picking and agricultural practices. 
Unlike existing datasets, such as CRAID \cite{Akiva2020craid}, which focuses on cranberry cultivation fields, \ourdataset encompasses a diverse range of wild berries (bilberries, cloudberries, crowberries, and lingonberries) captured in challenging conditions of forest undercanopies.
\ourdataset features 3,516 images with 18,336 annotated bounding boxes, \ourdataset provides a rich resource for developing and testing advanced object detection algorithms. 
We comprehensively analysed six popular object detectors (Faster R-CNN, VarifocalNet, GLIP, DINO, ObjectBox, and YOLOv8) to assess the dataset's utility in enhancing the performance and generalisation ability of detection models across varied forest regions and camera types.
One limitation of our dataset is the annotations for the bilberry species; \ie, we annotated bilberries and bog bilberries as the same class (species). 
These two species are very similar to each other, and only experts can accurately distinguish between them. 
As future work, one can explore domain adaptation techniques to improve cross-dataset transfer learning \cite{Mekhalfi2023}.
Moreover, we plan to involve experts for such fine-grained annotation to add more value to \ourdataset.

\vspace{2mm}
\noindent\textbf{Acknowledgement.} 
This work was supported by the EU Horizon Europe project FEROX under Grant n.~101070440.

%
%
\bibliographystyle{splncs04}
\bibliography{refs}
\end{document}